\documentclass[letterpaper, 10 pt, conference]{ieeeconf}

\IEEEoverridecommandlockouts
\usepackage{graphics}
\usepackage{epsfig}
\usepackage{times}
\usepackage{amsmath}
\usepackage{stfloats}
\usepackage{amssymb}
\usepackage{multirow}
\usepackage{subcaption}
\usepackage{booktabs}
\usepackage{float}
\usepackage{placeins}
\usepackage{amsfonts}
\usepackage{caption}

\usepackage{graphicx}
\usepackage[dvipsnames,table]{xcolor}
\usepackage{algorithmic}
\usepackage{paralist}
\usepackage{balance}

\newcommand{\plainimg}[2]{\includegraphics[width=#1]{#2}}

\newcounter{warpalgorithm}
\renewcommand{\thewarpalgorithm}{\arabic{warpalgorithm}}

\definecolor{srcgreen}{RGB}{0,112,60}
\definecolor{shiftred}{RGB}{176,32,32}
\newcommand{\taskpanel}[5]{%
  \begin{subfigure}[t]{0.241\textwidth}
    \centering
    \includegraphics[width=\linewidth]{#1}
    \caption{\textbf{#2}\\[1pt]
             #3\\[1pt]
             \textcolor{srcgreen}{#4\%}\,$\rightarrow$\,\textcolor{shiftred}{#5\%}}
  \end{subfigure}}

\title{\LARGE \bf
Warp: Structured Output-Space Adaptation for Frozen Robot Policies
}

\author{
    \authorblockN{Ethan Hirschowitz}
    \authorblockA{
        The University of Sydney, Australia \\
        \textit{ehir9923@uni.sydney.edu.au}
    }
    \and
    \authorblockN{Fabio Ramos}
    \authorblockA{
        The University of Sydney, Australia\\
        NVIDIA, USA\\
        \textit{fabio.ramos@sydney.edu.au}
    }
}

\begin{document}

\maketitle
\thispagestyle{empty}
\pagestyle{empty}

\begin{abstract}
Robot policies that remain broadly competent after deployment can degrade when dynamics, contact, geometry, actuation, or sensing differ from source conditions. We study post-hoc adaptation through \emph{structured output-space adapters}: compact transformations of samples from a frozen base policy. We introduce \emph{Warp}, a context-conditioned monotone bijective spline that enables nonlinear sample-dependent reshaping while preserving per-dimension ordering; Translation and Affine provide location-only and location--scale references. Across four ManiSkill tasks with controlled deployment shifts, Warp achieves the strongest recovery in a matched evolution-strategy (ES) adapter study, outperforming controls for spline parameterization, location--scale adaptation, and generic nonlinear capacity. Under PPO, the ordering becomes task dependent, showing that richer adapter structure does not automatically translate into better adaptation under a different optimizer. Finally, we test the practical benefit of this approach on physical SO101 peg insertion. Within an interaction-limited RLPD/SAC pipeline, Warp achieves comparable success to a strong stochastic residual while reducing median successful cycle time by approximately 30\%. These results identify structured output-space adaptation as a useful design space for adapting frozen robot policies.
\end{abstract}

\section{Introduction}
\label{sec:introduction}

Robot policies that perform reliably under source conditions can degrade after deployment when dynamics, contact, object geometry, actuation, calibration, or sensing change. Sim-to-real transfer is a particularly important instance: a simulation-trained policy may remain broadly competent on hardware while becoming systematically miscalibrated by effects that were absent or imperfectly modeled in simulation. Relearning the task from scratch discards useful behavior; a more targeted objective is to preserve the source policy and adapt how its actions are expressed after deployment. This setting is especially relevant when the deployed policy still contains useful task structure and only a comparatively small correction is required to recover effective behavior.

Residual policy learning is a prominent example of this strategy, but it raises a more general design question: \emph{what structure should an output-space adapter have?} A state-dependent translation can reposition a base action without changing the relative geometry of samples from the same state. An affine transform can additionally rescale those samples. A nonlinear sample-conditioned transform can change local geometry across the action distribution itself. These classes therefore differ not only in what corrections they can represent, but also in the parameterization presented to the learning algorithm. Choosing the adapter therefore determines which properties of the pretrained action distribution can be preserved, modified, or reshaped during adaptation.

\begin{figure*}[!t]
    \centering
    \includegraphics[width=0.82\linewidth]{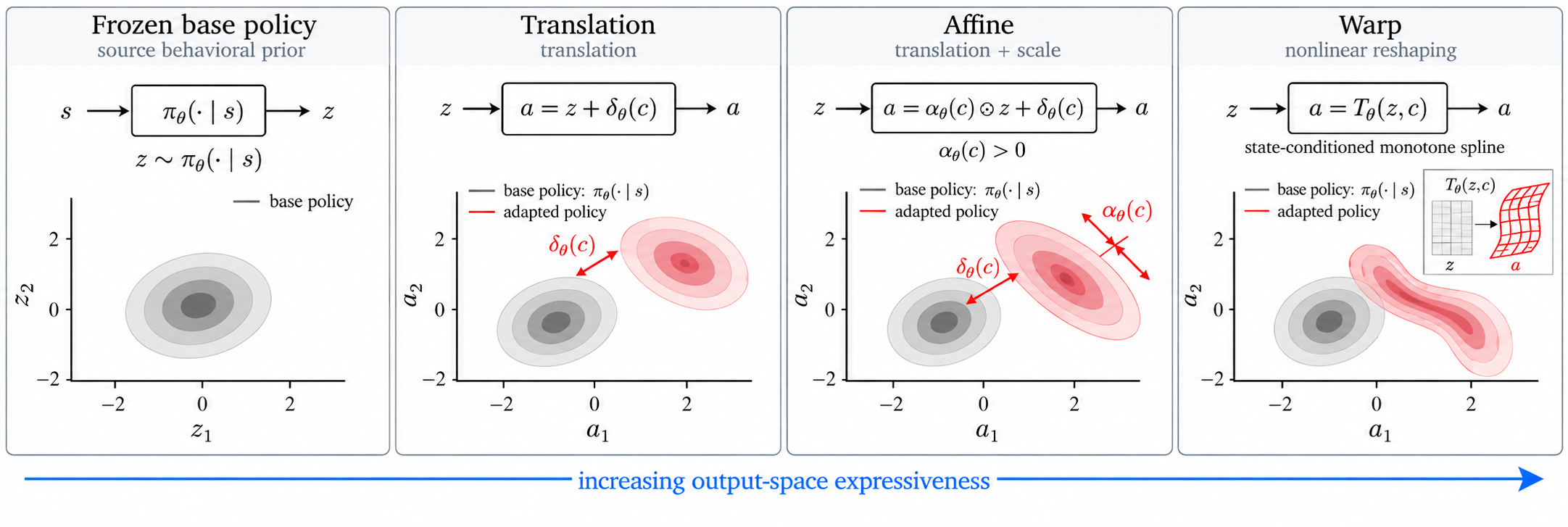}
    \caption{\textbf{Structured output-space adaptation of a frozen policy.} A frozen base policy defines the source action distribution. Translation changes action location, Affine additionally changes scale, and Warp applies a context-conditioned monotone bijection that enables nonlinear reshaping while preserving per-dimension ordering. The adapter classes therefore expose increasing output-space expressiveness while leaving the base policy fixed.}
    \label{fig:concept}
\end{figure*}

We study this design space as \emph{structured output-space adaptation}. Given a frozen stochastic base policy,
\begin{equation}
    z \sim \pi_{\mathrm{b}}(\cdot \mid s),
\end{equation}
we learn a compact transformation
\begin{equation}
    a = T_{\theta}(z,s)
    \label{eq:intro_adapter}
\end{equation}
while leaving the source policy fixed. This separates the behavioral prior encoded by the pretrained policy from the representation used to adapt it. Translation changes action location, Affine adds state-dependent scale, and nonlinear sample-conditioned transformations add sample-dependent local geometry. Fig.~\ref{fig:concept} summarizes the setting and the resulting adapter design space.

We instantiate the nonlinear class as \emph{Warp}, a context-conditioned monotone rational-quadratic spline applied to sampled base actions. Each per-dimension map is strictly monotone and bijective on its transform interval: ordering is preserved, identity initialization recovers the pretrained behavior, and the analytic inverse and Jacobian provide tractable densities. Warp is therefore more expressive than a location--scale correction while remaining substantially more structured than an unconstrained sample-conditioned network.

We evaluate Warp across matched gradient-free ES, on-policy PPO, and physical off-policy RLPD/SAC adaptation. The matched ES study provides the clearest architectural comparison, while PPO reveals task-dependent optimizer sensitivity; direct distributional analysis verifies non-affine reshaping. On physical SO101 peg insertion, Warp achieves comparable success to a strong stochastic residual while reducing median successful cycle time by approximately 30

Our contributions are:
\begin{compactitem}
    \item We formulate frozen-policy adaptation as a structured output-space transformation problem and introduce \emph{Warp}, an identity-initialized, bijective monotone spline adapter that enables nonlinear sampled-action reshaping with tractable densities.
    \item A matched ES study with Translation, SplineRes, Affine, and MLP controls isolates the benefit of structured warping; complementary PPO/fine-tuning comparisons and a shape-only distributional diagnostic characterize when that extra capability is used and when it improves return.
    \item We demonstrate the same structured-adaptation principle on physical SO101 sim-to-real peg insertion inside an interaction-limited RLPD/SAC pipeline against a strong stochastic residual baseline.
\end{compactitem}

\section{Related Work}
\label{sec:related_work}

\textbf{Residual and post-hoc robot adaptation.}
Residual policy learning improves an existing controller or policy by learning a corrective action on top of its nominal output. Early work combined model-free reinforcement learning with conventional feedback or model-based controllers when a useful but imperfect controller was already available~\cite{Johannink2018-uj,Silver2018-hz}. Recent methods extend this idea to learned robot policies. ResiP augments a frozen action-chunked behavior-cloning policy with a closed-loop PPO residual for precise assembly~\cite{Ankile2024-kq}, while residual off-policy RL learns lightweight per-step corrections to black-box behavior-cloning bases using sample-efficient off-policy updates~\cite{Ankile2025-ct}. Residual adaptation has also been extended to stochastic Gaussian and diffusion bases through base-action conditioning and uncertainty-guided exploration~\cite{Dodeja2025-up}. Together, these works show that residual adaptation spans different base policies, conditioning choices, stochasticity, and optimization algorithms.

\textbf{Sim-to-real and deployment-time adaptation.}
Sim-to-real transfer is a central instance of the deployment-shift problem considered here: the reality gap between simulated and physical systems can degrade a policy after transfer through mismatches in dynamics, sensing, calibration, contact, or actuation~\cite{Zhao2020-gk}. Domain and dynamics randomization attempt to anticipate such variation during source training by exposing the policy to simulated visual or physical changes~\cite{Tobin2017-ds,Peng2017-ku}. Other approaches build adaptation into the source policy itself, for example by conditioning control on online system identification~\cite{Yu2017-wv}, training an adaptation module for rapidly changing dynamics~\cite{Kumar2021-gv}, or explicitly meta-learning for fast target-task adaptation~\cite{Finn2017-li}. Post-hoc methods instead preserve a trained source policy and adapt after deployment. SPARR is particularly close to this practical regime: it retains a simulation-trained state-based assembly policy and learns a vision-conditioned stochastic residual from real-world interaction~\cite{Guo2026-hs}. We likewise focus on post-hoc adaptation of a competent frozen policy, but ask specifically how the \emph{structure} of the output-space transformation affects what can be corrected and how that correction is optimized.

\textbf{Policy fine-tuning and parameter-efficient adaptation.}
Direct fine-tuning instead updates the pretrained policy itself, and is a natural alternative to residual or output-space adaptation in robot learning~\cite{Alakuijala2021-vt,Yang2026SERNF}. Parameter-efficient methods provide a broader analogue: LoRA, for example, freezes pretrained weights while learning compact low-rank updates~\cite{Hu2021-lz}. Our approach pursues the same high-level goal of preserving a useful pretrained model, but localizes adaptation to the policy output rather than its parameters. This motivates our comparison against both full-policy and last-layer fine-tuning, which test whether modifying the source representation itself is preferable to retaining it and adapting only the action mapping.

\textbf{Expressive policy distributions and normalizing flows.}
Normalizing flows construct expressive distributions through invertible transformations of simple base densities, with rational-quadratic splines providing flexible monotone maps with analytic inverses and Jacobians~\cite{Papamakarios2019-he,Durkan2019-gm}. In RL, flows have primarily been used as the policy class itself: flow-based SAC variants replace Gaussian actors with richer exploration distributions~\cite{Ward2019-jf,Mazoure2020FlowsOffPolicy}, constrained flow policies encode action-space safety structure~\cite{Rietz2024-et}, and recent work studies flows as general-purpose models for policies and other RL distributions~\cite{Ghugare2025-lc}. SERNF further uses a likelihood-based normalizing-flow actor for sample-efficient real-world fine-tuning of action-chunked dexterous policies~\cite{Yang2026SERNF}. Warp differs in where the flow is placed: rather than replacing or fine-tuning the pretrained policy distribution, we retain it as the source distribution and learn only a compact, identity-initialized, state-conditioned transform of its sampled actions.

\section{Structured Output-Space Adaptation}
\label{sec:method}

\subsection{Problem Setting: Post-Hoc Adaptation of a Frozen Policy}
\label{sec:method_problem}

We consider a robot policy trained under source conditions and adapted after deployment without modifying its parameters. In simulation, source and shifted conditions share the task and action space while differing in physical properties such as object inertia or contact geometry. In sim-to-real transfer, deployment may additionally change sensing, calibration, actuation, and unmodeled hardware effects. Let the frozen stochastic base policy produce
\begin{equation}
    z \sim \pi_{\mathrm{b}}(\cdot\mid s),
\end{equation}
and let \(c\) denote the \emph{adapter context}, comprising the
deployment-time information available to the adapter at the current
timestep. The adapted action is
\begin{equation}
    a = T_{\theta}(z,c).
    \label{eq:general_adapter}
\end{equation}
In our state-based simulation experiments, \(c=s\), so the adapter is
state-conditioned. In the physical experiment, \(c\) additionally
includes the real-world observation and base action used by the
correction module. This notation lets the same formulation cover both
settings while keeping \(z\), the sampled base action being transformed,
distinct from the information \(c\) used to determine how it is
transformed.

The adapted policy is the pushforward of the base action distribution through \(T_\theta\). We optimize only the adapter parameters to maximize return under the deployment condition,
\begin{equation}
    \theta^{\star}
    =
    \arg\max_{\theta}
    \mathbb{E}_{\tau\sim\pi_\theta}
    \left[
        \sum_{t=0}^{H}\gamma^t R_t
    \right],
    \label{eq:adapt_objective}
\end{equation}
and initialize the adapter as a no-op,
\begin{equation}
    T_{\theta_0}(z,c)=z,
    \label{eq:identity_requirement}
\end{equation}
whenever the downstream action interface permits exact identity. The specific action-bounding and composition conventions used by each optimizer are reported with the corresponding experiment. The framework therefore specifies \emph{where} adaptation occurs---at the policy output---without tying the representation to a particular learning algorithm.

\subsection{Adapter Structure as a Capability Ladder}
\label{sec:method_hierarchy}

Within Eq.~\eqref{eq:general_adapter}, we distinguish adapters by how they can change the conditional action distribution. Translation applies the same state-dependent displacement to every sampled base action,
\begin{equation}
    T_{\mathrm{trans}}(z,c)=z+\delta_{\theta}(c),
    \label{eq:translation}
\end{equation}
so for a Gaussian base it moves the conditional mean while preserving source scale. We use this as a controlled \emph{Translation} reference rather than a proxy for all residual-RL methods. Affine adaptation adds positive per-dimension scaling,
\begin{equation}
    T_{\mathrm{aff}}(z,c)=\alpha_{\theta}(c)\odot z+\delta_{\theta}(c),
    \qquad \alpha_{\theta}(c)>0,
    \label{eq:affine}
\end{equation}
and can therefore change both location and scale. A nonlinear sample-conditioned adapter,
\begin{equation}
    T_{\mathrm{nonlin}}(z,c)=f_{\theta}(z;c),
    \label{eq:nonlinear_general}
\end{equation}
can additionally give two samples from the same base distribution different local corrections. Translation, Affine, and Warp thus form a capability ladder from location, to location--scale, to sample-dependent local reshaping. Table~\ref{tab:capability_ladder} summarizes these conceptual capabilities; the practical PPO implementations below are compared as methods rather than treated as a strictly nested ablation.

\begin{table}[t]
\centering
\caption{\textbf{Output-space adapter capability ladder.} The abstract classes differ in which aspects of the conditional action distribution they can modify.}
\label{tab:capability_ladder}
\footnotesize
\setlength{\tabcolsep}{5.0pt}
\renewcommand{\arraystretch}{1.05}
\begin{tabular}{lccc}
\toprule
\textbf{Adapter} & \textbf{Location} & \textbf{Scale} & \textbf{Local geometry}\\
\midrule
Translation & \checkmark & -- & --\\
Affine & \checkmark & \checkmark & --\\
Warp & \checkmark & \checkmark & \checkmark\\
\bottomrule
\end{tabular}
\end{table}

\subsection{Local Geometry Through the Adapter Jacobian}
\label{sec:method_jacobian}
The distinction is explicit in the Jacobian with respect to the sampled base action:
\begin{equation}
\begin{aligned}
J_{\mathrm{trans}}(z,c)&=I, \\
J_{\mathrm{aff}}(z,c)&=\operatorname{diag}(\alpha_{\theta}(c)), \\
J_{\mathrm{warp}}(z,c)&=\operatorname{diag}\!\left(\frac{\partial T_1}{\partial z_1},\ldots,\frac{\partial T_d}{\partial z_d}\right).
\end{aligned}
\label{eq:adapter_jacobians}
\end{equation}
Translation preserves local geometry, Affine introduces state-dependent but sample-independent scaling, and Warp allows the local scale to vary with both context and sampled action. It can therefore compress one part of a conditional marginal while expanding another without changing per-dimension ordering.

\subsection{Warp: Context-Conditioned Monotone Splines}
\label{sec:method_warp}

Warp instantiates the nonlinear transform with an elementwise monotone rational-quadratic (RQ) spline~\cite{Durkan2019-gm}. For action dimension \(i\),
\begin{equation}
    a_i=S_{\phi_i(c)}(z_i),
    \label{eq:spline_per_dim}
\end{equation}
where a conditioning network predicts the spline parameters \(\phi_i(c)\). With \(K\) bins, these consist of \(K\) positive bin widths, \(K\) positive bin heights, and \(K+1\) positive knot derivatives. Widths and heights are normalized to the transform interval and knot derivatives are constrained positive, making each one-dimensional map strictly monotone, bijective on that interval, and analytically invertible. The resulting elementwise transform is therefore bijective on the product action interval.

This parameterization imposes a deliberate inductive bias. Context conditioning lets the transformation vary across deployment-time inputs, while sample conditioning lets the correction vary across actions drawn under the same context. Monotonicity preserves per-dimension action ordering, so the adapter reshapes the base policy rather than arbitrarily permuting its action samples. The elementwise construction yields a diagonal Jacobian and computation linear in action dimension for a fixed number of spline bins. Its corresponding limitation is that the adapter does not introduce new instantaneous cross-dimensional dependence beyond that already encoded by the state-conditioned base policy.

\subsection{Identity Initialization and Tractable Density}
\label{sec:method_density}

Uniform spline bins and unit knot derivatives initialize Warp exactly to identity,
\begin{equation}
    S_{\theta_0}(z;c)=z,
    \qquad
    \log\left|\frac{\partial S_{\theta_0}}{\partial z}\right|=0.
    \label{eq:spline_identity}
\end{equation}
For any differentiable bijective adapter, change of variables gives, for $a=T_\theta(z,c)$,
\begin{equation}
    \log\pi_{\theta}(a\mid s)
    =\log\pi_{\mathrm{b}}(z\mid s)
    -\log\left|\det J_T(z,c)\right|,
    \label{eq:cov_logdensity}
\end{equation}
with $\log|\det J_T|=\sum_i\log|\partial T_i/\partial z_i|$ for the elementwise spline. Thus nonlinear reshaping retains an exact tractable likelihood.

\subsection{Bounded Continuous Actions}
\label{sec:method_bounded}

Robot commands are bounded, so we separate the learned bijection from any fixed action-domain map. In the canonical PPO implementation,
\begin{equation}
    u\sim\mathcal{N}(\mu_{\mathrm{b}}(s),\operatorname{diag}(\sigma_{\mathrm{b}}^2(s))),
    \quad z=\tanh(u),
    \quad a=S_{\theta}(z;s).
    \label{eq:bounded_warp}
\end{equation}
The pre-environment log density is
\begin{align}
    \log\pi_{\theta}(a\mid s)
    ={}&\log\mathcal{N}(u;\mu_{\mathrm{b}},\sigma_{\mathrm{b}})
    -\sum_i\log|1-\tanh^2(u_i)| \\
    &-\sum_i\log\left|\frac{\partial a_i}{\partial z_i}\right|.
    \label{eq:bounded_logprob}
\end{align}
The final term is Warp's Jacobian contribution; optimizer-specific clipping and composition conventions are reported with the corresponding experiments.

\subsection{Optimization Across Learning Regimes}
\label{sec:method_optimization}

The adapter defines the correction representation; the optimizer fits its parameters. PPO~\cite{Schulman2017-by} uses the exact density in Eqs.~\eqref{eq:cov_logdensity}--\eqref{eq:bounded_logprob}. Under evolution strategies (ES)~\cite{Salimans2017-yh}, black-box perturbations estimate
\begin{equation}
    \nabla_{\theta}J(\theta)
    \approx
    \frac{1}{n\sigma}
    \sum_{i=1}^{n}F_i\epsilon_i,
    \qquad \epsilon_i\sim\mathcal{N}(0,I),
    \label{eq:es_gradient}
\end{equation}
where $F_i$ is rollout return under $\theta+\sigma\epsilon_i$. The same bijective transform can also enter maximum-entropy off-policy objectives such as SAC through its log density. Holding the optimizer fixed still leaves different adapter parameterizations with different search spaces; we therefore use ES as the controlled structure study and PPO/off-policy learning as complementary practical regimes. The framework is not intrinsically Gaussian: ES requires only samples from the frozen base, while likelihood-based PPO/SAC require a tractable base density so that the transformed density can be evaluated. Algorithm~\ref{alg:warp} summarizes the common adapter placement and optimizer-specific update path.

\refstepcounter{warpalgorithm}\label{alg:warp}
\begin{center}
\begin{minipage}{0.98\columnwidth}
\footnotesize
\hrule
\vspace{2pt}
\textbf{Algorithm \thewarpalgorithm: Adapt a frozen policy with Warp}
\vspace{2pt}
\hrule
\vspace{1pt}
\begin{algorithmic}[1]
\STATE \textbf{Input:} frozen base policy $\pi_{\mathrm b}$; identity-initialized Warp $T_{\theta_0}$ (Eq.~\eqref{eq:spline_identity}); deployment environment; optimizer $\mathcal A$
\FOR{adaptation iterations}
  \STATE Observe $s$ and form adapter context $c$ from available deployment-time inputs ($c=s$ in simulation)
  \STATE Sample $z\sim\pi_{\mathrm b}(\cdot\mid s)$; keep $\pi_{\mathrm b}$ fixed
  \STATE Execute $a=T_\theta(z,c)$ using the bijective spline in Eq.~\eqref{eq:spline_per_dim}
  \STATE Store the resulting deployment transition
  \IF{$\mathcal A$ is likelihood-based (PPO/SAC)}
    \STATE Evaluate $\log\pi_\theta(a\mid s)$ via Eqs.~\eqref{eq:cov_logdensity}--\eqref{eq:bounded_logprob} and update only $\theta$
  \ELSE
    \STATE Evaluate parameter perturbations and update $\theta$ with ES (Eq.~\eqref{eq:es_gradient})
  \ENDIF
\ENDFOR
\STATE \textbf{Return:} frozen $\pi_{\mathrm b}$ with adapted output transform $T_{\theta^\star}$
\end{algorithmic}
\vspace{1pt}
\hrule
\end{minipage}
\end{center}

\section{Experiments}
\label{sec:experiments}

We ask four questions. \textbf{(Q1)} Under a shared optimizer, how does adapter structure affect recovery after a deployment shift? \textbf{(Q2)} Does that ordering persist under PPO and policy fine-tuning? \textbf{(Q3)} Do trained Warp adapters actually use the nonlinear reshaping that distinguishes them from location--scale adaptation? \textbf{(Q4)} Does structured warping remain useful in a practical physical sim-to-real pipeline? We study these questions with matched ES, PPO and fine-tuning, a direct distributional diagnostic, and off-policy RLPD/SAC on hardware.

\subsection*{Common Simulation Setup}
\label{sec:exp_common}

\begin{figure*}[h]
  \centering
  \taskpanel{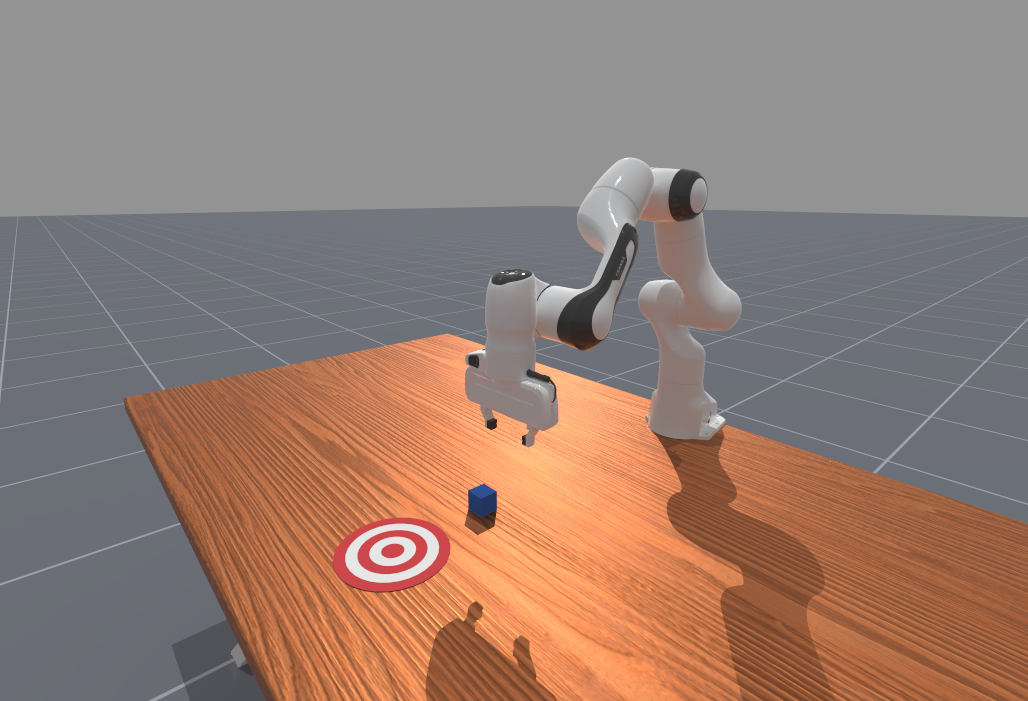}{PushCube}{mass $\times 500$}{87.6}{13.4}\hfill
  \taskpanel{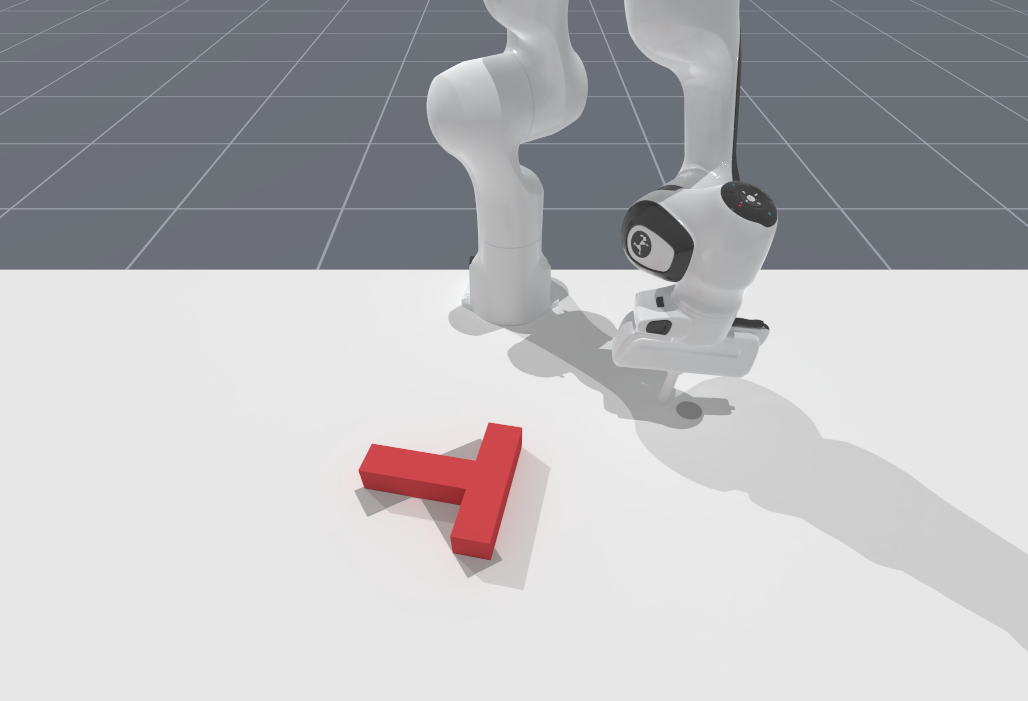}{PushT}{mass $\times 5$}{98.8}{31.6}\hfill
  \taskpanel{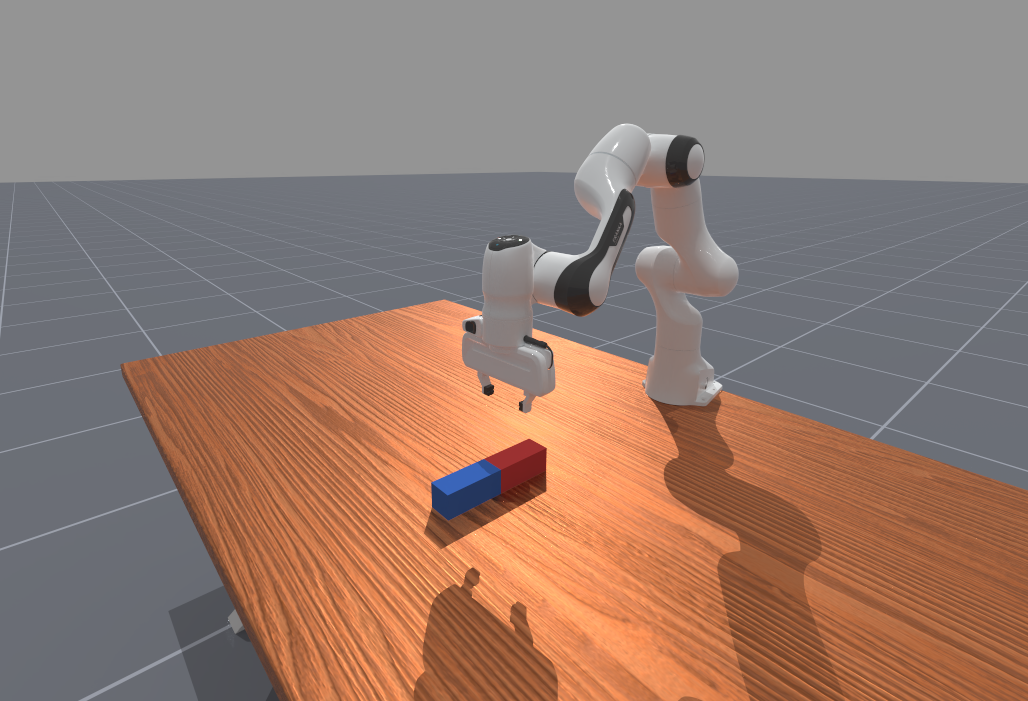}{LiftPegUpright}{mass $\times 2$}{76.2}{32.4}\hfill
  \taskpanel{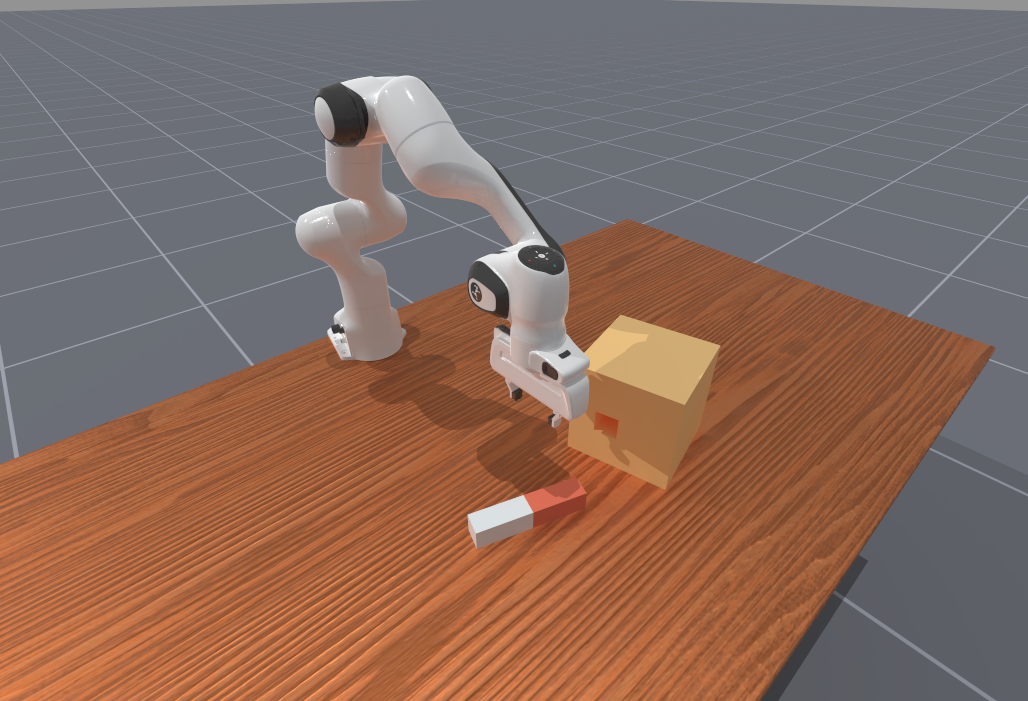}{PegInsertionSide}{clearance $\times 0.5$}{86.6}{59.6}
  \caption{\textbf{Controlled deployment shifts in ManiSkill3.} Each panel reports the task, the deployment shift, and frozen-base success from the source condition (\textcolor{srcgreen}{green}) to the shifted condition (\textcolor{shiftred}{red}). The suite contains three inertial shifts and one contact-geometry shift; adapted-policy results are reported in Tables~\ref{tab:es_ablation} and~\ref{tab:ppo_natural}.}
  \label{fig:tasks}
\end{figure*}

We use four ManiSkill3 manipulation tasks~\cite{Tao2024-cz}: PushCube, PushT, LiftPegUpright, and PegInsertionSide. A Gaussian PPO policy is trained under source conditions and then frozen. We introduce one controlled deployment shift per task: PushCube mass \(\times 500\), PushT mass \(\times 5\), LiftPegUpright mass \(\times 2\), and PegInsertionSide clearance \(\times 0.5\). The first three are inertial shifts; the fourth changes contact geometry. PushCube is comparatively insensitive to smaller mass changes, so the larger factor is used to create a meaningful frozen-base performance gap. Fig.~\ref{fig:tasks} shows the tasks and the corresponding frozen-base performance drops. Observations are proprioceptive state vectors and actions are bounded continuous commands.

Within each task, all adapters start from the same source checkpoint and receive the same interaction budget. Periodic simulation evaluation uses 32 held-out environments; we retain the checkpoint with highest monitored return and report it on a separate \(5\) training-seed \(\times\,5\) evaluation-seed grid, so checkpoint selection and final reporting are seed-disjoint. All ES adapters use a common ES optimizer and rollout budget. For the PPO comparison, methods share the shifted condition, source checkpoint, rollout budget, and PPO hyperparameters. Following recommended practice for deep-RL evaluation~\cite{Henderson2018DRLMatters,Agarwal2021Precipice}, we report IQM with 95\% bootstrap confidence intervals. Simulation correction networks use two hidden layers of width 256. Spline-based adapters use \(K=4\) bins except on PegInsertionSide, where \(K=8\). Main simulation adaptation runs use 2M environment steps; PPO uses learning rate \(3\times10^{-4}\), while ES uses a population of 1024 and perturbation scale \(\sigma=0.05\).

\subsection*{Q1: How Does Adapter Structure Affect Gradient-Free Adaptation?}
\label{sec:exp_es}

\begin{table*}[!b]
\centering
\caption{\textbf{Matched ES study of adapter structure.} Entries are IQM success (\%) over five training seeds. The final row reports aggregate recovery \(\rho\) with 95\% bootstrap confidence intervals across task--seed runs.}
\label{tab:es_ablation}
\small
\setlength{\tabcolsep}{6pt}
\renewcommand{\arraystretch}{1.08}
\begin{tabular}{lccccc}
\toprule
\textbf{Task} & \textbf{Translation} & \textbf{SplineRes} & \textbf{Affine} & \textbf{MLP} & \textbf{Warp}\\
\midrule
PushCube & 96.4 & 94.1 & 83.8 & 84.8 & \textbf{99.4}\\
PushT & 32.3 & 23.2 & 15.0 & 18.2 & \textbf{69.3}\\
LiftPegUpright & 32.2 & 92.0 & 91.3 & 92.0 & \textbf{98.6}\\
PegInsertionSide & 63.3 & 67.4 & 47.9 & 47.1 & \textbf{71.0}\\
\midrule
Aggregate \(\rho\) & 0.21 [0.08, 0.38] & 0.68 [0.59, 0.73] & 0.34 [0.31, 0.37] & 0.38 [0.35, 0.41] & \textbf{0.88 [0.81, 0.93]}\\
\bottomrule
\end{tabular}
\end{table*}

We use ES for the controlled adapter-structure study because the compact correction modules can be optimized directly from rollout returns without policy likelihoods. The controls target distinct explanations: \emph{Translation} provides a location-only reference; \emph{SplineRes} uses the same spline machinery, conditioning network, parameter count, initialization, and ES optimizer as Warp, but evaluates the spline at a fixed input to produce only a state-dependent additive offset; \emph{Affine} adds state-dependent scale; an unconstrained sample-conditioned \emph{MLP} tests generic nonlinear access to the base sample; and \emph{Warp} passes the sampled base action through the structured monotone spline.

To aggregate recovery across tasks, we define
\begin{equation}
    \rho =
    \frac{J_{\mathrm{method}}-J_{\mathrm{shift}}}
         {J_{\mathrm{source}}-J_{\mathrm{shift}}},
    \label{eq:recovery_ratio}
\end{equation}
where \(J_{\mathrm{shift}}\) is the frozen-base return after the deployment shift, so \(\rho=0\) denotes no recovery and \(\rho=1\) denotes full recovery of source-condition performance. Table~\ref{tab:es_ablation} reports task-level success and aggregate recovery. Warp is strongest on every task and reaches aggregate recovery \(0.88\,[0.81,0.93]\), compared with \(0.68\,[0.59,0.73]\) for SplineRes, \(0.38\) for the unconstrained sample-conditioned MLP, \(0.34\) for Affine, and \(0.21\) for Translation. The non-overlapping aggregate confidence intervals make the SplineRes--Warp comparison particularly informative. The controls identify two distinct effects. First, spline parameterization itself produces a substantially better ES search space: SplineRes outperforms the conventional Translation network in aggregate. Second, directly transforming the sampled base action provides an additional benefit beyond parameterization alone, with Warp exceeding SplineRes on all four tasks. Generic sample-conditioned nonlinearity is insufficient, as the MLP remains far below Warp. The result therefore supports structured warping under matched ES rather than an advantage from optimizer choice or parameter count alone. The task-level pattern is also informative. PushT is the clearest structured-Warp case: the shifted base is at 31.6\%, SplineRes falls to 23.2\%, and Warp reaches 69.3\%. PushCube is near ceiling for both Translation and Warp, consistent with a simpler location correction being largely sufficient. On LiftPegUpright, several richer controls recover most of the source performance but Warp remains highest; on PegInsertionSide, the separation is smaller, with SplineRes approaching Warp. Together, these cases show that the value of structured nonlinear reshaping is shift dependent even within the matched ES setting.

\subsection*{Q2: Does the ES Ordering Persist under PPO?}
\label{sec:exp_natural}

Under PPO, \emph{Translation} learns a state-dependent mean offset while retaining the frozen exploration scale; \emph{Affine} additionally learns positive state-dependent scale in the Gaussian latent space; and canonical bounded \emph{Warp} applies the state-conditioned RQ spline after the base sample is mapped into the bounded action domain. We also compare full-policy KL-regularized fine-tuning (\emph{FT-KL}, \(\beta=0.1\)) and last-layer fine-tuning (\emph{FT-Last}). Because Translation/Affine and canonical Warp act in different transformation domains, this is a practical method comparison rather than a strictly nested capacity ablation.

Table~\ref{tab:ppo_natural} shows that the ES ordering does not persist. Warp has the highest output-space adapter point estimate on PushCube and LiftPegUpright, whereas Affine is highest on PushT and PegInsertionSide; uncertainty overlaps substantially on the latter comparisons. Affine also fails to consistently improve over Translation, and fine-tuning is competitive on individual tasks without dominating the suite. Thus richer adapter capacity does not automatically improve PPO adaptation.

\begin{table*}[!t]
\centering
\caption{\textbf{PPO adaptation under controlled deployment shifts.} Shifted Base is the frozen source policy evaluated after the deployment shift. Learned methods report IQM success (\%) with 95\% bootstrap confidence intervals over five training seeds and five held-out evaluation seeds; boldface marks the largest point estimate in each task.}
\label{tab:ppo_natural}
\small
\setlength{\tabcolsep}{4.1pt}
\renewcommand{\arraystretch}{1.08}
\begin{tabular}{lcccccc}
\toprule
\textbf{Task} & \textbf{Shifted Base} & \textbf{Translation} & \textbf{Affine} & \textbf{Warp} & \textbf{FT-KL} & \textbf{FT-Last}\\
\midrule
PushCube & 13.4 & 95.0 [93.6, 97.3] & 91.0 [86.6, 95.6] & \textbf{97.5 [96.1, 99.3]} & 74.1 [69.1, 81.4] & 96.6 [93.5, 98.1] \\
PushT & 31.6 & 54.5 [48.9, 61.4] & 57.4 [55.7, 59.8] & 51.5 [49.6, 52.9] & \textbf{63.8 [60.7, 66.9]} & 60.1 [55.0, 68.8] \\
LiftPegUpright & 32.4 & 75.0 [69.3, 85.3] & 81.6 [73.1, 88.9] & \textbf{87.3 [76.5, 92.5]} & 87.2 [84.1, 89.1] & 71.0 [62.7, 85.2] \\
PegInsertionSide & 59.6 & 57.4 [55.1, 59.2] & \textbf{63.1 [55.3, 69.8]} & 57.8 [54.9, 61.8] & 60.7 [55.7, 63.9] & 58.8 [55.7, 59.6] \\
\bottomrule
\end{tabular}
\end{table*}

\subsection*{Q3: Does Warp Actually Use Nonlinear Reshaping?}
\label{sec:exp_distribution}

To test whether trained Warp policies actually use their nonlinear capability, we analyze one-dimensional action marginals at fixed held-out states. Pre- and post-Warp samples are independently standardized before computing Wasserstein-1 distance~\cite{PeyreCuturi2019OT}; translation and positive affine scaling vanish under this normalization in population, so the remaining distance measures non-affine shape change. Identity, translation, and affine controls provide numerical null references.

\begin{figure}[!b]
    \centering
    \includegraphics[width=\linewidth]{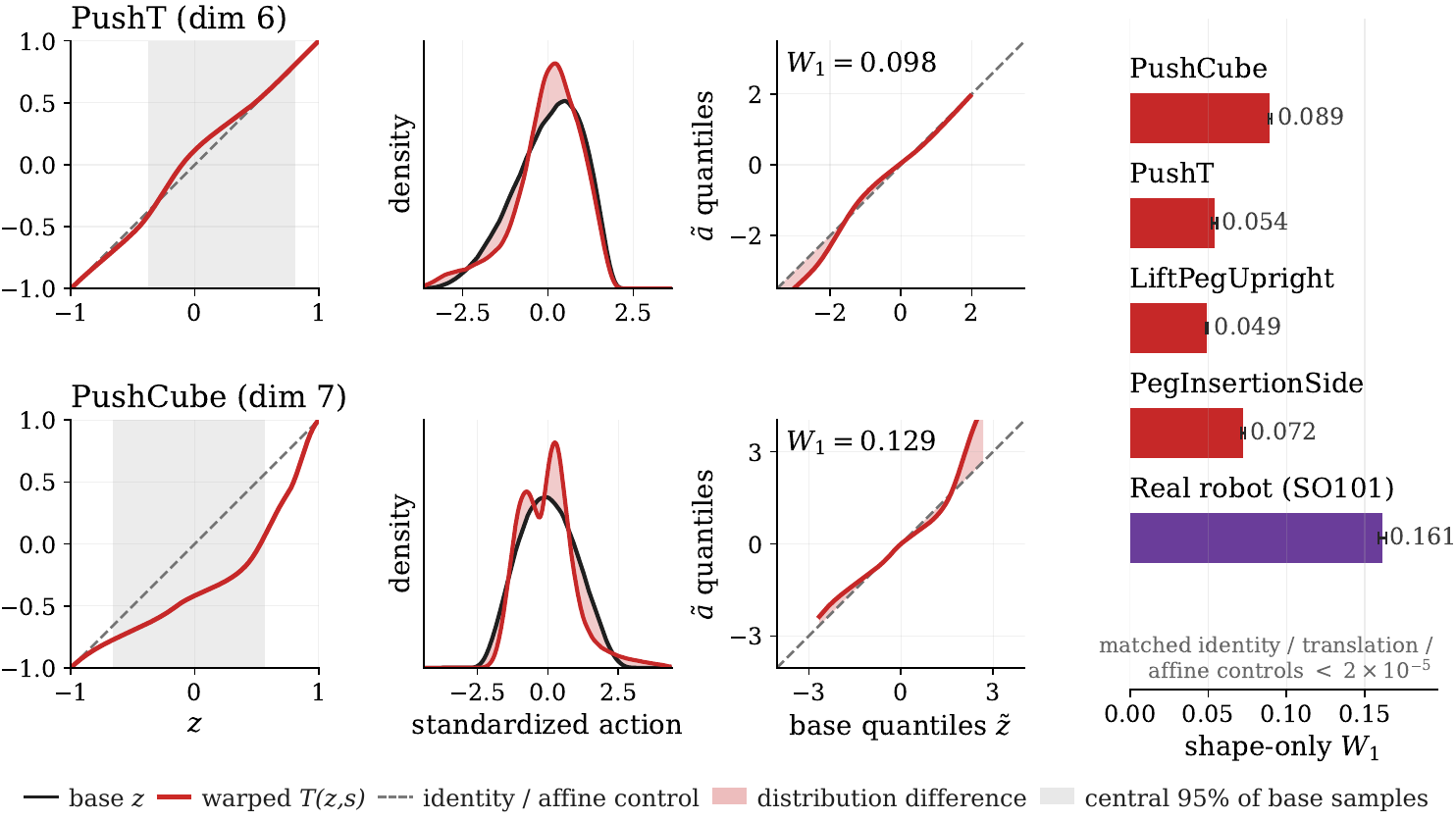}
    \caption{\textbf{Warp uses non-affine reshaping.} Learned spline maps and standardized marginals show shape changes beyond translation or positive affine scaling; shape-only Wasserstein distance summarizes the effect across simulation tasks and SO101.}
    \label{fig:reshaping}
\end{figure}

Simulation tasks yield aggregate shape-only \(W_1\) between \(0.049\) and \(0.089\), while the physical SO101 policy yields \(0.161\). Matched identity/translation/affine controls remain below \(2\times10^{-5}\), with finite-sample null values around \(0.017\)--\(0.023\). Fig.~\ref{fig:reshaping} shows representative transforms, standardized marginals, and the aggregate diagnostic. Warp therefore makes measurable non-affine changes rather than collapsing to a complicated affine correction. This is a representation-level diagnostic rather than a performance metric: it establishes that the nonlinear capability is used, not that greater shape change must improve return. The analysis is intentionally marginal; because Warp is elementwise, it does not establish new instantaneous cross-dimensional dependence.

\subsection*{Q4: Does Warp Transfer to Practical Sim-to-Real Adaptation?}
\label{sec:exp_robot}

\begin{figure}[h]
  \centering
  \setlength{\tabcolsep}{1.5pt}
  \renewcommand{\arraystretch}{1}
  \begin{tabular}{@{}c@{\hspace{2pt}}ccc@{}}
    \rotatebox[origin=c]{90}{\footnotesize\textit{Sim}} &
      \plainimg{0.30\linewidth}{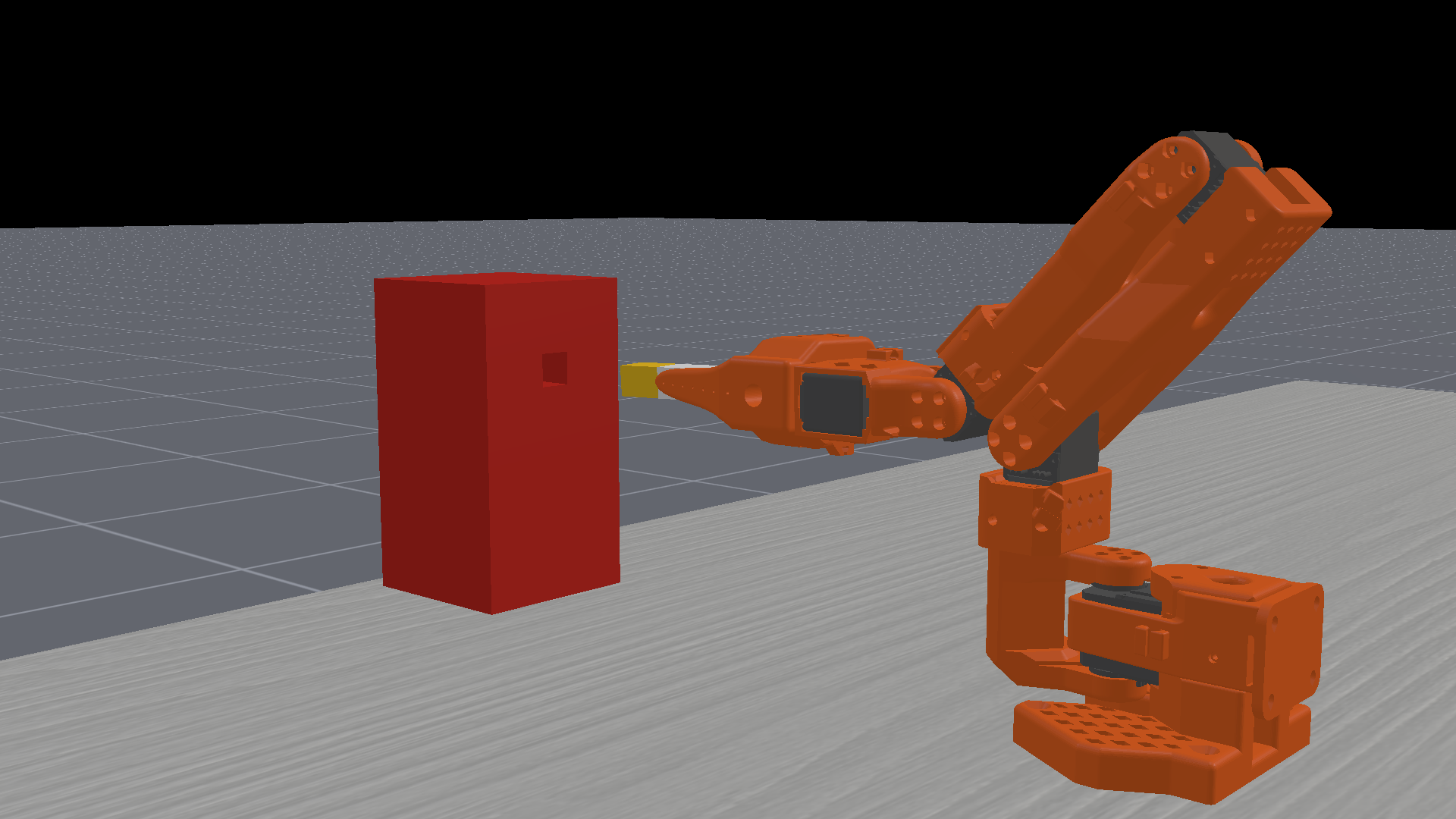} &
      \plainimg{0.30\linewidth}{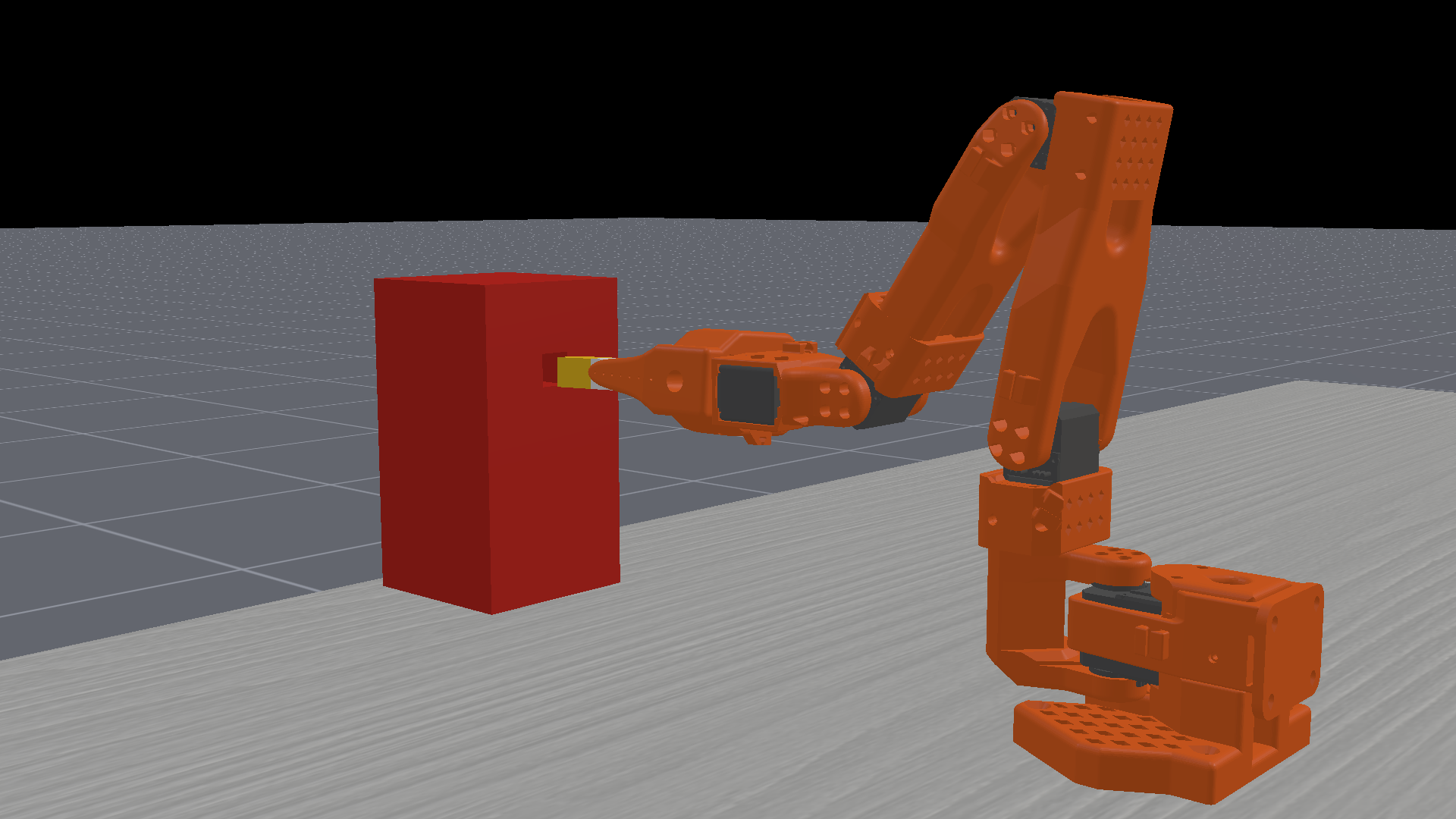} &
      \plainimg{0.30\linewidth}{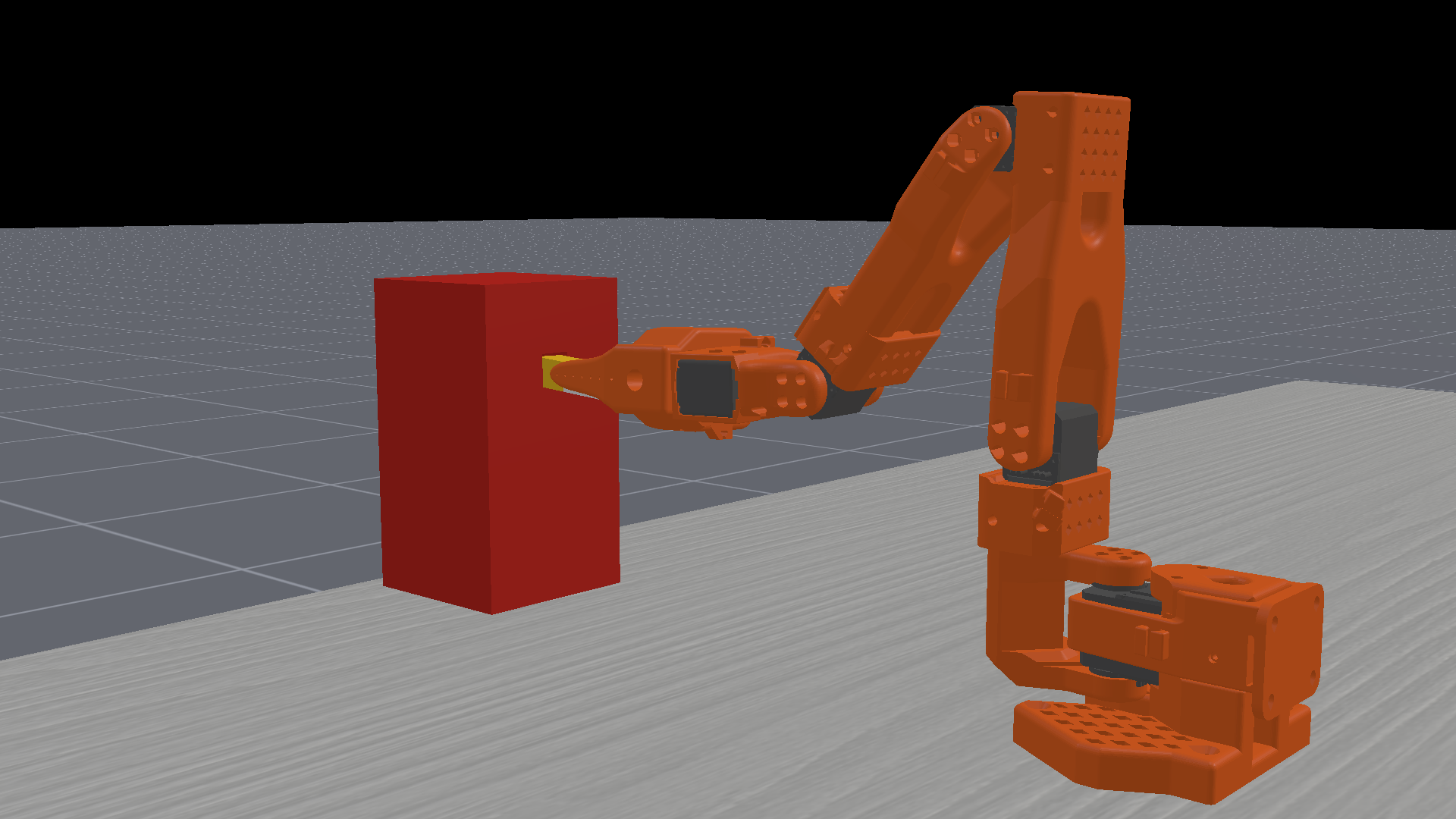} \\[1pt]
    \rotatebox[origin=c]{90}{\footnotesize\textit{Real}} &
      \plainimg{0.30\linewidth}{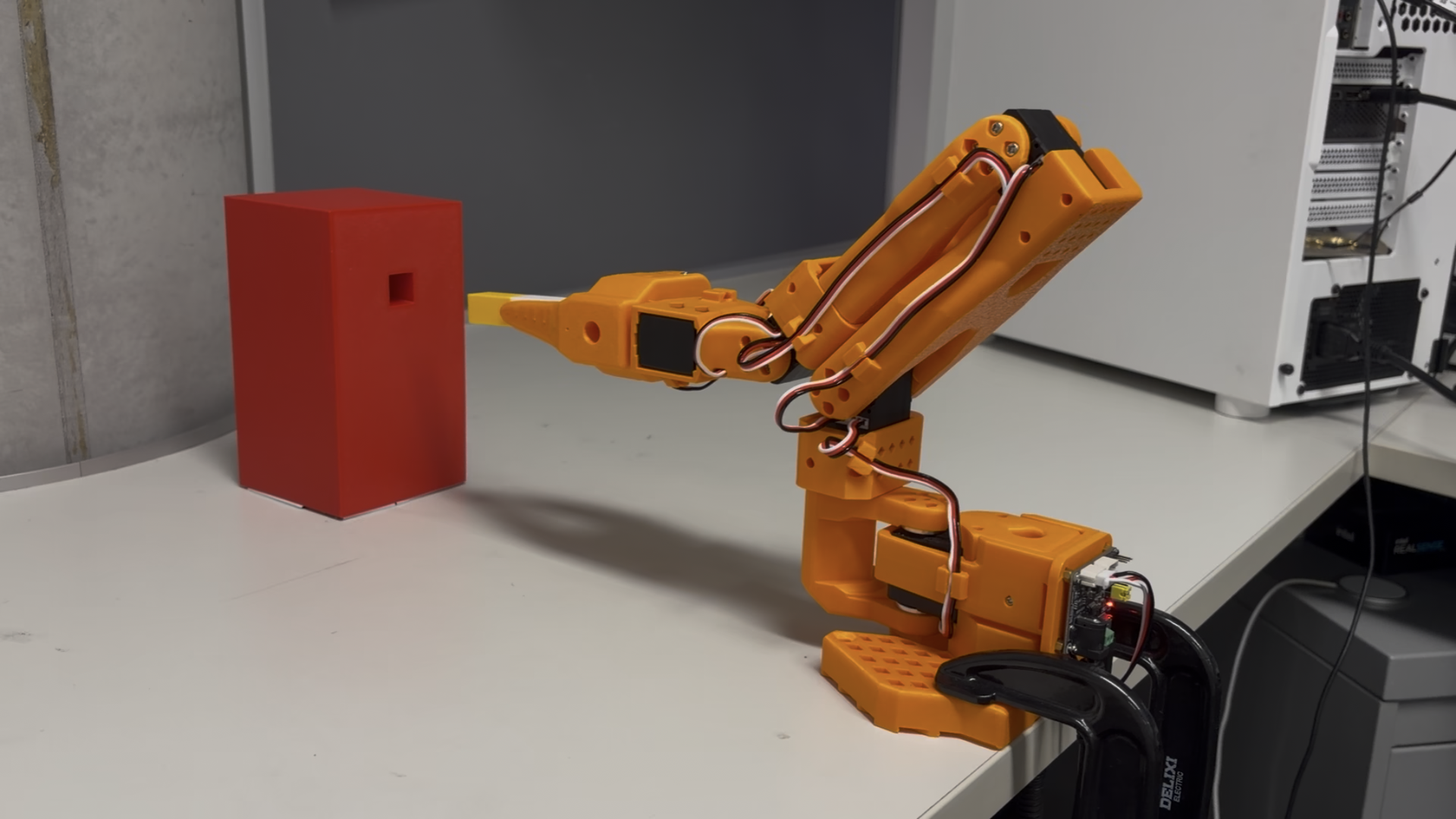} &
      \plainimg{0.30\linewidth}{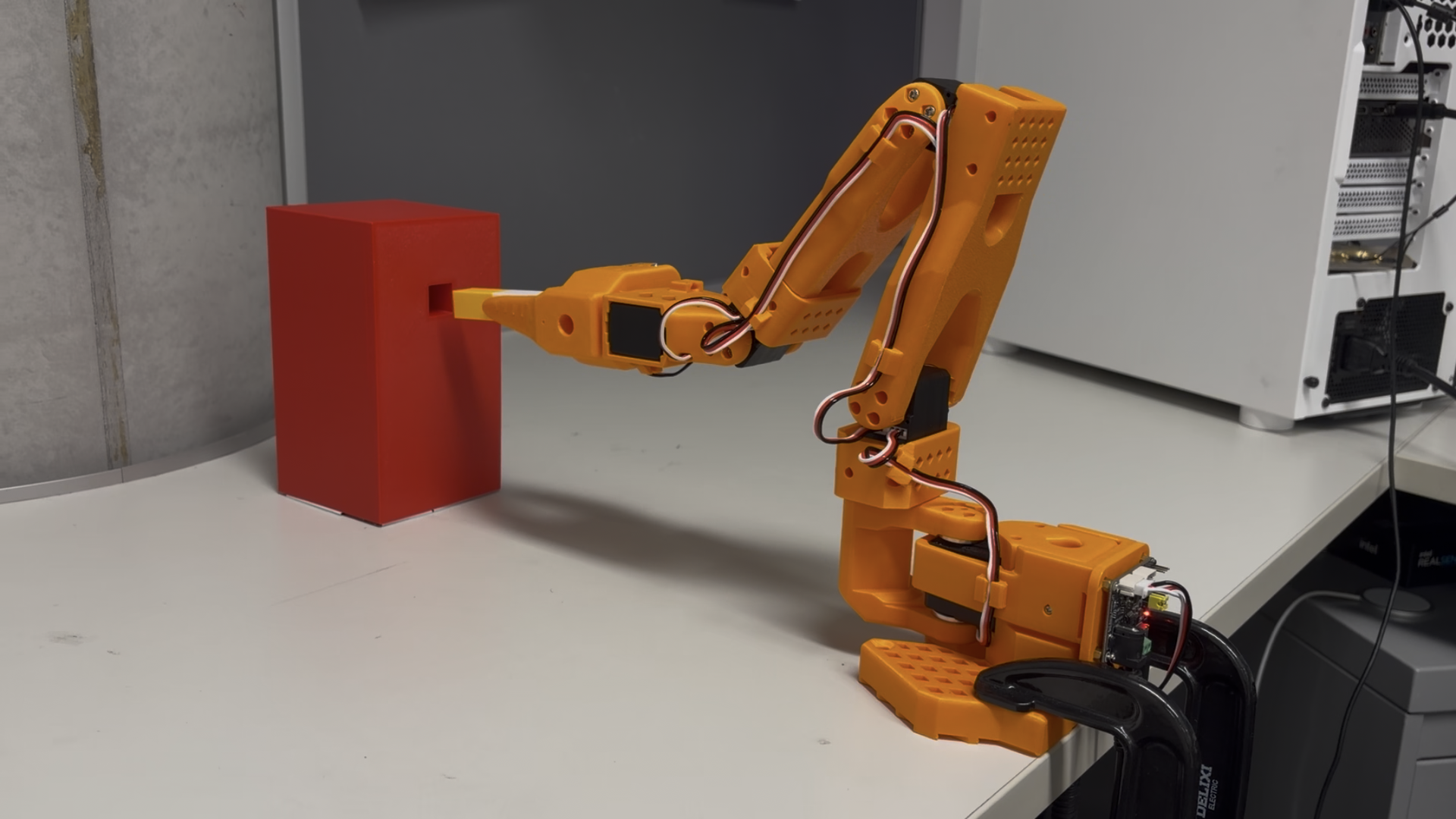} &
      \plainimg{0.30\linewidth}{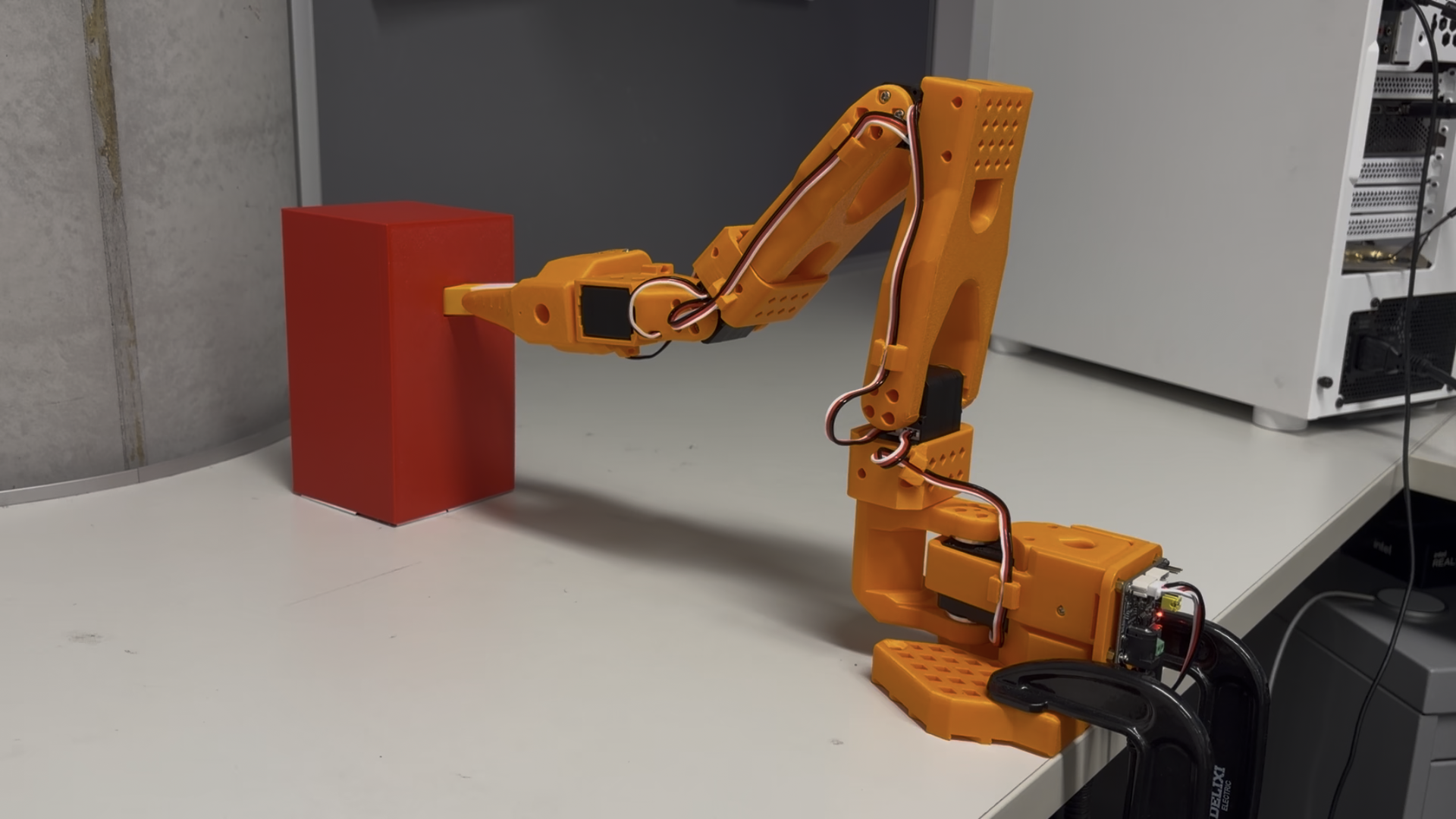} \\[2pt]
     & \footnotesize Start & \footnotesize Contact & \footnotesize Inserted \\
  \end{tabular}
  \caption{\textbf{SO101 sim-to-real peg insertion.} Simulation (top) and physical robot (bottom) are shown at three phases of a successful episode. The PPO base policy is trained in simulation and frozen before transfer; residual and Warp corrections are learned only from real-world interaction.}
  \label{fig:so101_sequence}
\end{figure}

We finally test Warp in a practical, interaction-limited off-policy sim-to-real adaptation stack. We benchmark Warp against a strong stochastic residual on physical SO101 peg insertion following a SPARR-style regime~\cite{Guo2026-hs}. The PPO base is trained on the corresponding simulated SO101 task and frozen before transfer. The peg--hole clearance is only 2\,mm. At this tolerance, geared-servo backlash and friction, calibration offsets, structural compliance, and imperfect contact modeling can compound at the end effector. The task is therefore semantically simple but mechanically stringent, making the SO101 platform a focused test of deployment-time mismatch. Fig.~\ref{fig:so101_sequence} shows the simulated and physical task at representative phases of a successful episode.

Twenty successful real-world trajectories seed replay, and the stochastic residual and Warp are each trained for 50 online robot episodes with the same RLPD~\cite{Ball2023-um}/SAC~\cite{Haarnoja2018-or} pipeline. The residual actor is stochastic, conditions on the real-world observation and base action, and learns its own exploration scale, making it substantially stronger than the Translation attribution reference used in simulation. Warp retains the same frozen base, conditioning, replay initialization, critic structure, controller, interaction budget, and action post-processing, but parameterizes the stochastic correction with the spline transform. Thus the comparison changes the correction parameterization while holding the surrounding adaptation pipeline fixed.

For deterministic evaluation, the residual executes its mean correction. Because \(T_\theta(\mathbb{E}[z])\neq \mathbb{E}[T_\theta(z)]\) in general, Warp approximates its expected transformed correction using 64 stochastic samples in one batched forward pass. Each method is then evaluated for 20 deterministic episodes. Table~\ref{tab:robot} reports success and successful-cycle-time statistics.

\begin{table}[t]
\centering
\caption{\textbf{Physical SO101 peg insertion.} Results use 20 deterministic evaluation episodes; cycle time is computed over successful episodes only.}
\label{tab:robot}
\footnotesize
\setlength{\tabcolsep}{3.5pt}
\renewcommand{\arraystretch}{1.08}
\begin{tabular}{lccc}
\toprule
\textbf{Method} & \textbf{Success} & \textbf{Mean (s)} & \textbf{Median (s)}\\
\midrule
Frozen base & 45\% & 4.73 \(\pm\) 0.27 & 4.80\\
Stochastic residual & 65\% & 4.66 \(\pm\) 0.19 & 4.65\\
Warp & 70\% & 3.64 \(\pm\) 0.93 & \textbf{3.25}\\
\bottomrule
\end{tabular}
\end{table}

Both learned corrections improve over the frozen base. Warp and the stochastic residual achieve comparable success at this evaluation scale (70\% versus 65\%), while Warp reduces median successful cycle time from 4.65\,s to 3.25\,s, approximately 30\%. With one task and 20 evaluation episodes, we treat the one-episode success difference cautiously and view execution efficiency as the clearer signal. For repeated successful assembly, lower cycle time translates directly into higher nominal task throughput, so this efficiency difference is operationally meaningful even without a resolved success-rate advantage. Together with the non-affine \(W_1\) observed for the physical Warp policy, the result shows that structured warping remains active inside an interaction-limited stochastic residual-learning stack.

\section{Discussion}
\label{sec:discussion}

The matched ES study provides the clearest evidence that output-space adapter structure matters. Warp is the only adapter that performs strongly across all four deployment shifts, while the SplineRes and MLP controls separate two effects: spline parameterization improves the ES search space, and directly applying the structured transform to sampled base actions provides a further benefit beyond state-only spline parameterization or generic sample-conditioned nonlinearity. This result supports Warp as a useful structured adapter under matched gradient-free optimization.

The PPO study places an important boundary on that conclusion. The ES ordering does not persist: Warp has the strongest output-space adapter point estimate on PushCube and LiftPegUpright, whereas Affine is highest on PushT and PegInsertionSide, with substantial uncertainty on the latter comparisons. Additional location--scale or nonlinear capacity therefore does not guarantee better PPO adaptation. We interpret the ES experiment as the controlled architectural study and the PPO experiment as evidence that the practical value of a representation depends on the optimizer used to fit it, rather than as a universal expressiveness hierarchy.

The distributional analysis clarifies what the mixed PPO results do and do not mean. Learned Warp policies exhibit non-affine changes in standardized action marginals across every simulation task and on the physical policy, so Warp is not simply collapsing to a complicated Translation or Affine adapter. Whether that extra reshaping improves return is a separate optimization question. The mechanism diagnostic therefore establishes that the representation is being used without claiming that nonlinear shape change must help on every task.

The hardware experiment provides a complementary practical test. Its baseline is a stronger stochastic, base-action-conditioned residual actor trained with the same RLPD/SAC pipeline and real-world interaction budget. Warp achieves comparable success at this evaluation scale while reducing median successful cycle time by approximately 30\%. We view this as evidence that structured sample transformation can be integrated into an interaction-limited off-policy adaptation stack, not as evidence of a statistically resolved success-rate advantage from a single 20-episode evaluation.

The present evidence also defines the scope of our conclusions. The simulation suite contains three inertial shifts and one contact-geometry shift rather than an exhaustive catalogue of deployment mismatch; friction, actuation, sensing, and broader contact changes remain important tests. The current Warp is elementwise and per-timestep, so it does not introduce new sampled cross-dimensional or temporal coupling. Finally, output-space adaptation assumes that the frozen source policy remains a useful behavioral prior: it is intended to refine competent behavior, not to recover a fundamentally incorrect source policy.

\section{Conclusion}
\label{sec:conclusion}

We cast post-hoc adaptation of frozen robot policies as a structured output-space transformation problem. Warp uses a bijective monotone spline to add nonlinear sample-dependent reshaping beyond Translation and Affine while preserving the source policy. Under matched ES, Warp achieves the strongest recovery across four deployment shifts and outperforms controls for affine scaling, spline parameterization, and generic sample-conditioned nonlinearity; under PPO, the ordering is task dependent. Distributional analysis verifies that Warp actually uses non-affine reshaping, and physical SO101 experiments show that the same principle can operate in a sample-efficient off-policy sim-to-real pipeline. Structured output-space adaptation therefore provides a practical middle ground between simple residual correction and full policy fine-tuning, with both representation and optimization shaping performance.

\section*{Acknowledgment}
Large language models (LLMs) were used as assistive tools for notation and consistency checks, and grammatical editing of this manuscript. LLMs were also used in the practical component of this work, for implementation support and experiment/run orchestration. Models were also used to assist with generating presentation-ready visualizations. All generated code, experiment configurations, outputs, analyses, and manuscript edits were reviewed and verified by the authors, who retain full responsibility for the work.

\balance
\bibliographystyle{IEEEtran}
\bibliography{paper}

\end{document}